\documentclass[journal,twoside,web]{ieeecolor}
\usepackage{jsen}
\usepackage{cite}
\usepackage{amsmath,amssymb,amsfonts}
\usepackage{algorithm}
\usepackage{algpseudocode}
\usepackage{graphicx}
\usepackage{textcomp}
\usepackage{wrapfig}
\usepackage{siunitx}

\makeatletter
\renewcommand{\ALG@beginalgorithmic}{\small}
\makeatother

\def\BibTeX{{\rm B\kern-.05em{\sc i\kern-.025em b}\kern-.08em
    T\kern-.1667em\lower.7ex\hbox{E}\kern-.125emX}}
\definecolor{abstractbg}{rgb}{0.89804,0.94510,0.83137}
\begin{document}
\title{Neuromorphic circuit for temporal odor encoding in turbulent environments}
\author{Shavika Rastogi, Nik Dennler, Michael Schmuker, and André van Schaik \IEEEmembership{Fellow, IEEE}
\thanks{Part of this work was funded by an NSF/MRC award under the Next Generation Networks for Neuroscience initiative (NeuroNex Odor to action, NSF \#2014217, MRC \#MR/T046759/1)
}
\thanks{
Affiliations: S.R., N.D., and A.v.S: International Centre for Neuromorphic Systems, Western Sydney University, Kingswood 2747 NSW, Australia. 
S.R., N.D., and M.S.: 
Biocomputation Group, University of Hertfordshire, Hatfield AL10 9AB, United Kingdom. 
M.S.: BioML Consulting, Berlin, Germany.
E-mail: rastogi.shavika@gmail.com; dennler@proton.me; m.schmuker@biomachinelearning.net; a.vanschaik@westernsydney.edu.au.
}}

\IEEEtitleabstractindextext{%
\fcolorbox{abstractbg}{abstractbg}{%
\begin{minipage}{\textwidth}%
\begin{wrapfigure}[12]{r}{2.7in}%
\includegraphics[width=2.5in]{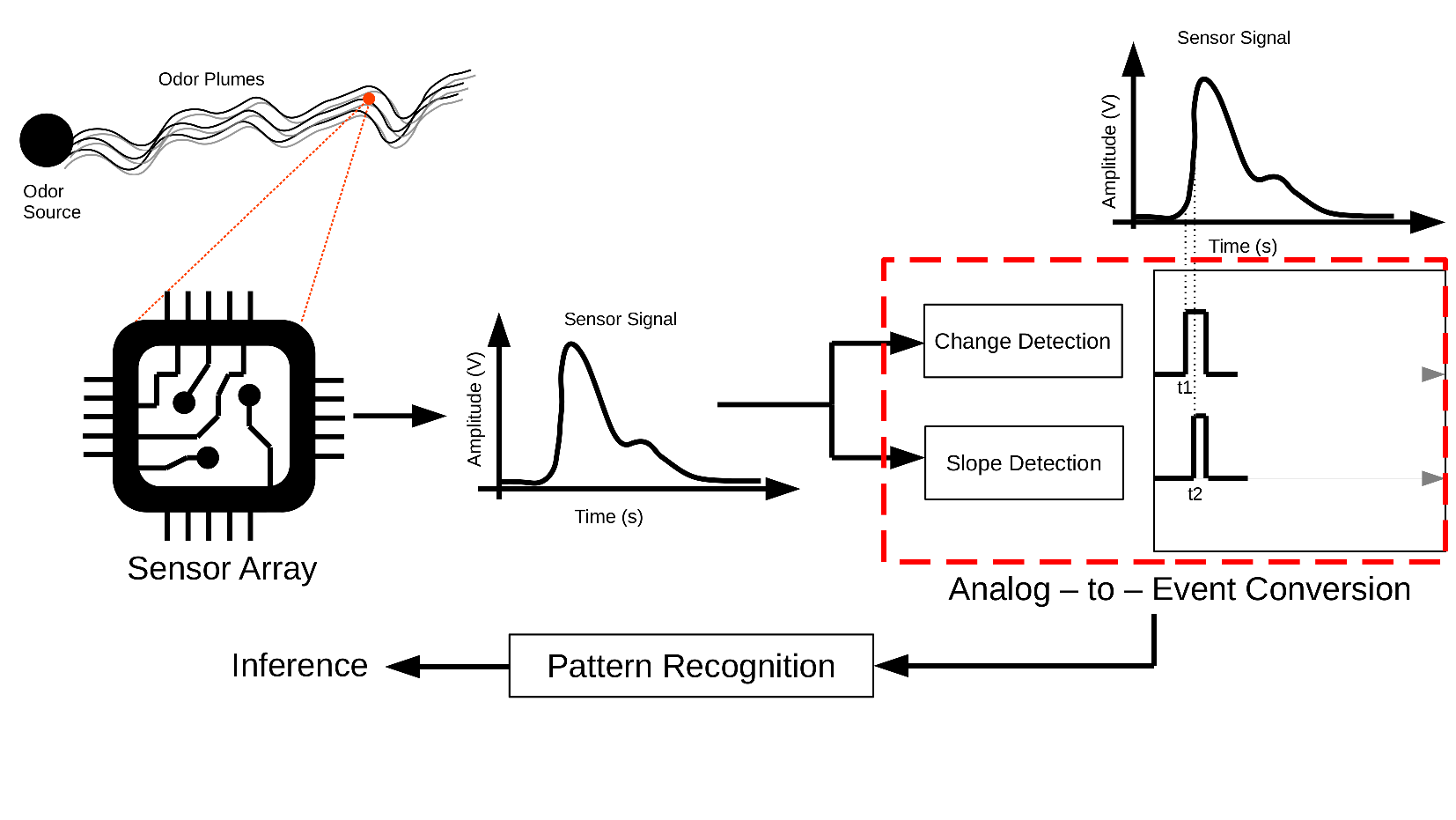}%
\end{wrapfigure}%
\begin{abstract}
Natural odor environments present turbulent and dynamic conditions, causing chemical signals to fluctuate in space, time, and intensity. 
While many species have evolved highly adaptive behavioral responses to such variability, the emerging field of neuromorphic olfaction continues to grapple with the challenge of efficiently sampling and identifying odors in real-time.
In this work, we investigate Metal-Oxide (MOx) gas sensor recordings of constant airflow-embedded artificial odor plumes. 
We discover a data feature that is representative of the presented odor stimulus at a certain concentration---irrespective of temporal variations caused by the plume dynamics. 
%
%
%
Further, we design a neuromorphic electronic nose front-end circuit for extracting and encoding this feature into analog spikes for gas detection and concentration estimation. 
The design is inspired by the spiking output of parallel neural pathways in the mammalian olfactory bulb. 
We test the circuit for gas recognition and concentration estimation in artificial environments, where either single gas pulses or pre-recorded odor plumes were deployed in a constant flow of air. 
For both environments, our results indicate that the gas concentration is encoded in---and inversely proportional to---the time difference of analog spikes emerging out of two parallel pathways, similar to the spiking output of a mammalian olfactory bulb.
The resulting neuromorphic nose could enable data-efficient, real-time robotic plume navigation systems, advancing the capabilities of odor source localization in applications such as environmental monitoring and search-and-rescue.  
\end{abstract}

\begin{IEEEkeywords}
Neuromorphic front-end, gas concentration, MOx sensors, machine olfaction, mammalian olfactory bulb

\end{IEEEkeywords}
\end{minipage}}}

\maketitle

\section{Introduction}
\label{sec:introduction}
%
\IEEEPARstart{O}{dors} are transported in the air by dynamic plumes---shaped by dispersion processes such as advection and diffusion \cite{Riffell2008}.
Within these odor plumes, gas concentration fluctuates over time and space, offering temporal features that are informative of the environment.
Insects utilize the temporal dynamics of odor plumes to locate plants and distant mates \cite{Conchou2019}, inspiring the development of computationally cheap and optimized algorithms for source localization in mobile robotics \cite{Chen2019} and other fast sensing applications.
For instance, performing automated measurement campaigns of odor concentration is crucial for monitoring industrial emissions, ensuring regulatory compliance, and mitigating the impact of harmful atmospheric pollutants \cite{Conti2020}.
Rapid gas concentration detection systems should be robust to the changes in odor plume dynamics for the accurate detection of gas and its concentration level in these scenarios. \par

The frequency of odor concentration fluctuations in an open and turbulent environment is correlated with the distance between source and sensor \cite{MYLNE1991, Yee1994, Celani2014,  Schmuker2016}. In such environments, the absolute gas concentration at any point is highly variable and lacks smooth gradients. This stresses the importance of sampling and measuring odor encounters at different points in time, eventually allowing for localizing and identifying the gas source \cite{Schoepe2024}. Neuromorphic methods \cite{Vanarse2017}\cite{Persaud2013} offer a data-efficient solution in such scenarios through asynchronous sampling \cite{Safa2022} and encoding these odor events in the form of spikes \cite{Rastogi2023}. In turbulent environments with multiple odor events, encoding gas concentration information in the timing of spikes rather than in spike frequencies may be advantageous, eliminating the need to average over multiple spike intervals for each event
\cite{Rullen2001}\cite{Sengupta2017}.   \par




In our previous work \cite{Rastogi2024}, we proposed neuromorphic analog front-ends for individual  Metal-Oxide (MOx) gas sensors 
and for MOx sensor arrays to encode the gas concentration level as a time difference between two generated spike events---similarly to the odor concentration encoding in the mammalian olfactory bulb \cite{Fukunaga2012}. However, the proposed front ends were designed and tested using data that reflect a highly controlled environment, with single gas pulses at a fixed concentration and pulse width. 
In this scenario, the response amplitude of a MOx sensor increases with rising gas concentrations and is indicative of a given concentration level of gas. 
It has been shown that MOx sensors can capture rapid temporal odor fluctuations with sub-second precision \cite{Drix2021, dennler2024}, despite their finite integration time.  
However, as in turbulent plumes both the concentration and duration of a stimulus will fluctuate, we expect the MOx sensor response to vary even for the same instantaneous absolute concentration.  
In this case, the amplitude of the MOx sensor response alone may not be a reliable indicator of the odor concentration.  
\par

The primary objective of this work is to analyze MOx sensor recordings of artificial odor plumes and to design a front-end circuit for MOx electronic nose (e-nose) system. This circuit aims to 
accurately encode gas concentration magnitude and gas identity into discrete events, irrespective of temporal variations in gas plume dynamics. 
Inspired by the mammalian olfactory bulb \cite{Fukunaga2012}, the circuit encodes gas concentration levels for each encounter through spike timings in two parallel pathways. Further, we will validate the circuit's performance using the single gas pulse scenario introduced in our previous work. \par 



\section{Materials and Methods}


For analyzing plume data and designing our circuit, we use a publicly available dataset \cite{dennler2024_dryad} that was collected with a high-speed electronic nose \cite{dennler2024}, consisting of sensor response curves with respect to different stimuli. The e-nose comprises eight MOx sensors, from which four were operated at a constant and closed-loop controlled sensor heater temperature, which allows for precise measurements of odor fluctuations. From these four sensors, we here use the response of one reducing (RED, Sensor 1) and one oxidizing (OX, Sensor 2) sensor, both embedded in the MiCS-6814 package. 

The experimental setup to deliver the odor stimuli has first been described in Ackels et al. \cite{Ackels2021}, and in a modified form been used in the data collection campaign \cite{dennler2024, dennler2024_dryad}. An olfactometer comprising multiple gas manifolds and a set of fast microvalves was used, where the valves were opened and closed at \SI{500}{\hertz}. By modulating the duty cycle, arbitrary lower-frequency odor stimuli can be produced. To ensure constant airflow across the odor stimuli, the odor valves were compensated by valves that deliver a non-odorous stimulus of the same magnitude.

In this work, we use the MOx sensor response curves that resulted from odor plumes replayed by the olfactometer. For this, turbulent plumes were recorded in an outdoor environment using a 2-channel Photoionization Detector (PID) \cite{Ackels2021}. Two odor sources were either placed in close proximity to each other or separated by \SI{50}{\centi\meter}, resulting in different degrees of correlation between the two incoming odors. Further, these PID traces were cropped to \SI{5}{\second} trials, normalized, and then used to shape the stimulus delivered by the olfactometer. In particular, the duty cycle of two odor valves and two compensation valves were modulated to reproducibly replicate the PID response, while ensuring constant flow across the stimulus. 

Different combinations of gas pairs in different orders were delivered using this method. From these, the e-nose response for the following pairs was used for our analysis: (1) Ethyl Butyrate and Eucalyptol (EB\textunderscore Eu and Eu\textunderscore EB) (2) Ethyl Butyrate and Isoamyl Acetate (EB\textunderscore IA and IA\textunderscore EB), and (3) Isoamyl Acetate and Eucalyptol (IA\textunderscore Eu and Eu\textunderscore IA).\par

\begin{figure}[h]
\includegraphics[width=\linewidth]{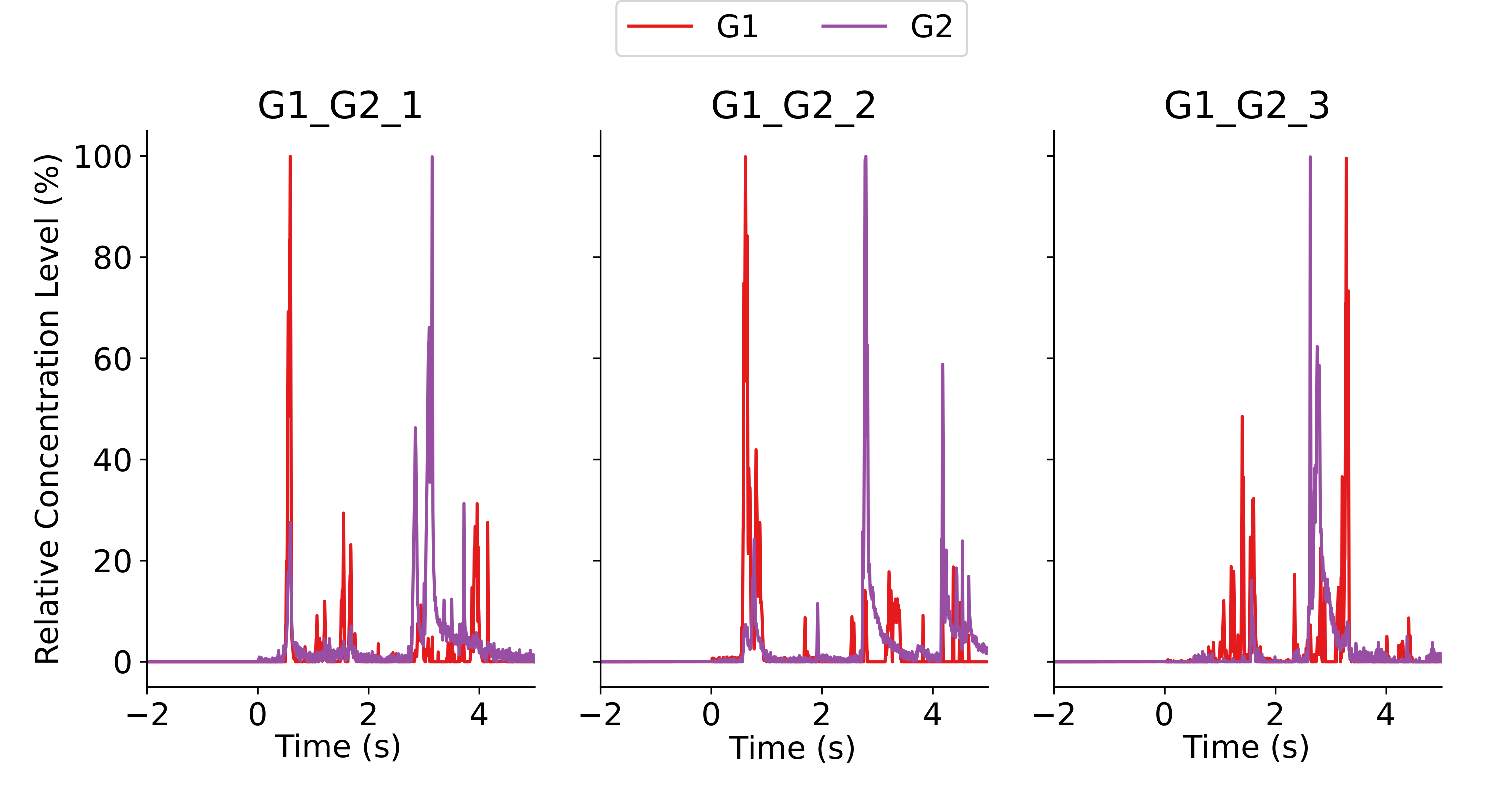}
\caption{PID traces of plumes used to control odor valves of gases G1 and G2. The y-axis depicts the gas concentration level measured by PID relative to the highest used gas concentration. The number associated with the label of each PID trace represents the trial number.}
\label{fig:plume_snapshots}
\end{figure}

Figure \ref{fig:plume_snapshots} shows three out of all PID traces that were used to control the odor valves of the setup. Each plume trace starts at $t=\SI{0}{\second}$ and has a duration of \SI{5}{\second}. Negative timestamps indicate the time before the start of the plume trace. The MOx data recorded in different gas orders for these three traces only have been used for our analysis. These three traces out of all available traces are chosen because the peaks of different gases are well separated. The y-axis indicates the relative gas concentration levels reflected in the PID measurements. The range of gas concentration levels is from C1 to C5, which is the same as in the single gas pulse environment described in our previous study \cite{Rastogi2024}. The PID recordings are such that the 20 \% gas concentration represents the lowest concentration level C1, 40 \% gas concentration represents the concentration level C2, and so on. In this way, 100 \% gas concentration represents the highest concentration level of C5. \par

\begin{figure}[h]
\includegraphics[width=\linewidth]{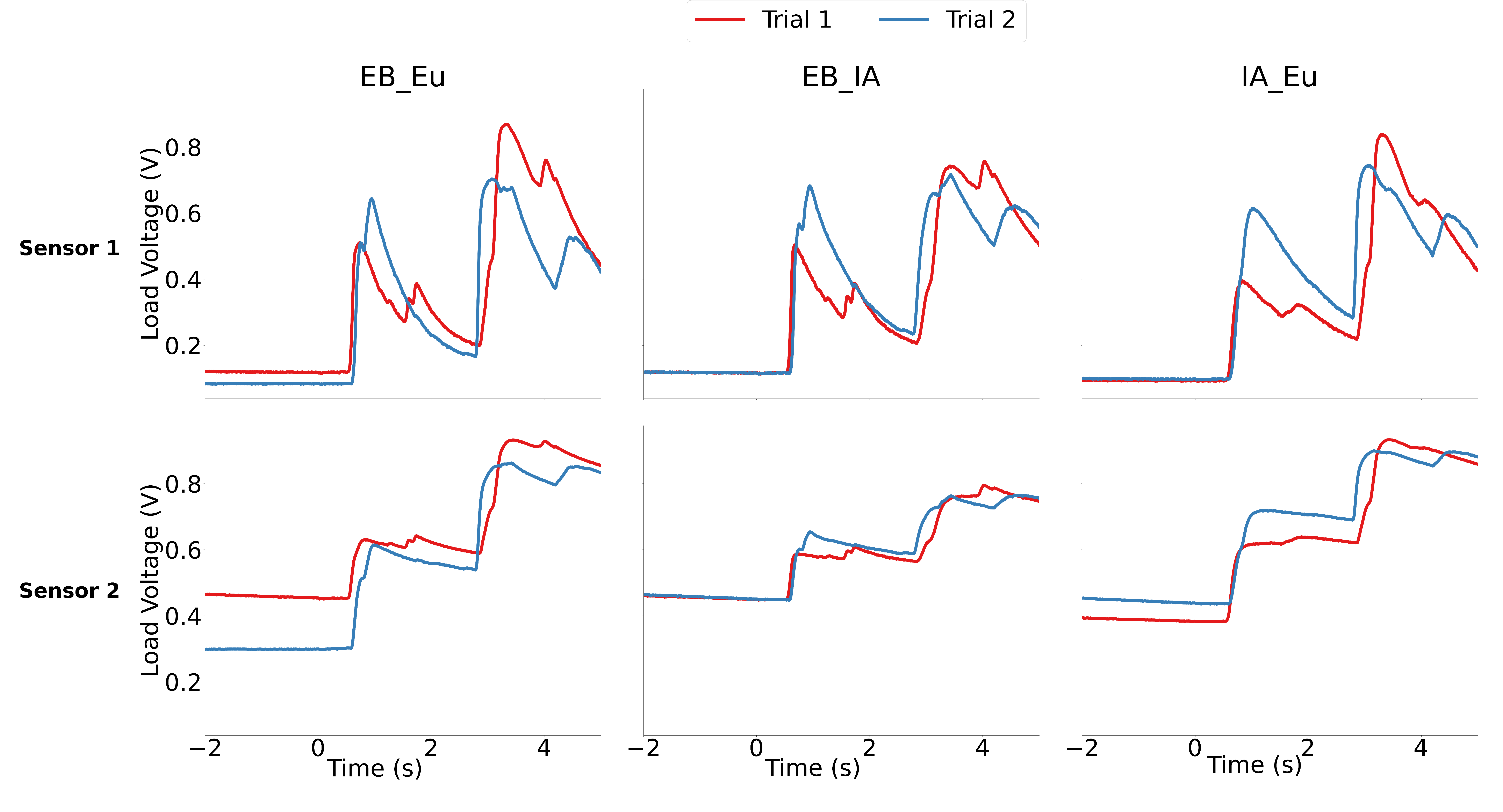}
\caption{MOx recordings for plume traces 1 and 2 for different gas combinations.}
\label{fig:mox_data_plumes_1_2}
\end{figure}

Figure 2 shows MOx recordings for plume traces 1 and 2 for different gas combinations. These two plume traces are different trials of the same plume as observed in Figure 1. It can be seen that there are some obvious differences in the MOx recordings for a given gas at the same concentration level in these two plume traces. This suggests a need to extract a feature from the MOx recordings that can represent the concentration level of a specific gas across different traces of the same plume. This feature should be robust to the variations in temporal dynamics of the gas.

\section{Analysis of MOx data for Plumes}

The two prominent peaks of MOx recordings shown in Figure 2 indicate the highest gas concentration level in these two plume traces. Visual inspection of these peaks shows that MOx sensors produce varying response amplitudes for the same gas concentration in two different traces of the same plume. The exposure measurement (EM) circuit used in our previous work \cite{Rastogi2024} integrates the area under the MOx signal curve, and its output is influenced by the amplitude variations in the MOx signal. Therefore, the EM circuit will not perform reliable concentration measurements for realistic plumes.  \par

The visual inspection of these two prominent peaks also reveals that for a gas at a given concentration level, the slope of the rising edge of the MOx signal appears to be similar in both plume traces. We will call the rising edge of the filtered MOx signal "bout" and its slope "bout slope". Therefore we checked whether the bout slope can be a robust indicator of gas identity and its concentration level in the plume environment irrespective of differences in the plume dynamics at different points of time. \par

\subsection{Bout Slope Analysis for Plume Dataset}

From the PID traces shown in Figure 1, it can be seen that the odor-background segregation is easier for the first two prominent peaks, where the gas of higher concentration is known as the target odor. Therefore, it is expected that the MOx sensor response for these two peaks indicates the presence of the dominant gas irrespective of the gas present in the background. In an alternative scenario, different gases may have similar concentration levels at a particular point in time. If the MOx sensors 
used in this environment have similar sensitivities to all gases, it would not be feasible to identify a specific gas and its concentration level at that time point using MOx recordings. For the PID traces used in this work, we are therefore only interested in the detection of the first two prominent peaks in the MOx recordings. These two peaks lie in the duration of 0-2s and 2-5s respectively in these traces. In the third PID trace, there are two consecutive peaks in the duration of 2-5s which are not well separated. Therefore, we will not be looking at the peaks in the duration of 2-5s in the MOx recordings for this trace.  

\begin{table}
\caption{Different trials of MOx recordings for each gas at two different peak positions.}
\centering
\label{table}
\setlength{\tabcolsep}{3pt}
\begin{tabular}{|p{25pt}|p{75pt}|p{75pt}|}
  \hline
  Gas & Peak 1 & Peak 2 \\
  \hline
  EB & EB\textunderscore Eu\textunderscore 1 \par
  EB\textunderscore Eu\textunderscore 2 \par
  EB\textunderscore IA\textunderscore 1 \par
  EB\textunderscore IA\textunderscore 2 &
  Eu\textunderscore EB\textunderscore 1 \par
  Eu\textunderscore EB\textunderscore 2 \par
  IA\textunderscore EB\textunderscore 1 \par
  IA\textunderscore EB\textunderscore 2 \\
  
  \hline
  Eu & Eu\textunderscore EB\textunderscore 1 \par
  Eu\textunderscore EB\textunderscore 2 \par
  Eu\textunderscore IA\textunderscore 1 \par
  Eu\textunderscore IA\textunderscore 2 &

  EB\textunderscore Eu\textunderscore 1\par
  EB\textunderscore Eu\textunderscore 2 \par
  IA\textunderscore Eu\textunderscore 1 \par
  IA\textunderscore Eu\textunderscore 2 \\
  \hline
  IA & IA\textunderscore Eu\textunderscore 1 \par
  IA\textunderscore Eu\textunderscore 2 \par
  IA\textunderscore EB\textunderscore 1 \par
  IA\textunderscore EB\textunderscore 2 &
  Eu\textunderscore IA\textunderscore 1 \par
  Eu\textunderscore IA\textunderscore 2 \par
  EB\textunderscore IA\textunderscore 1 \par
  EB\textunderscore IA\textunderscore 2 \\
  \hline
    \end{tabular}
\label{tab1}
\end{table}

\begin{table}
\caption{Plume Recordings for each gas where the gas concentration level is lower than the highest concentration level at the first peak.}
\centering
\label{table}
\setlength{\tabcolsep}{3pt}
\begin{tabular}{|p{25pt}|p{75pt}|}
\hline
Gas & Plume Recordings \\
\hline
EB & EB\textunderscore Eu\textunderscore 3 \par
EB\textunderscore IA\textunderscore 3 \\

\hline
Eu & Eu\textunderscore EB\textunderscore 3 \par
Eu\textunderscore IA\textunderscore 3 \\

\hline

IA & IA\textunderscore EB\textunderscore 3 \par
IA\textunderscore Eu\textunderscore 3 \\

\hline

\end{tabular}
\label{tab2}
\end{table}

Table 1 shows the trials of plume recordings for each gas depending on whether that particular gas is present in the first or in the second dominant peak of the recording. It contains only those trials in which these two peaks represent the highest concentration level of any gas. It can be seen that there are four such recordings for each gas at each peak position. The analysis will be done only on the largest peaks in the duration 0-2s and 2-5s in the plume traces mentioned in Table 1. \par

Table 2 shows the trials of plume recordings for each gas where the gas concentration is lower than the highest concentration level at the first peak. The concentration level is possibly between C2 and C3 for all gases in the duration of 0-2s. We have two trials for each gas at this concentration level. The analysis will be done only on the largest peak in the duration 0-2s in these traces.\par



To compute the slope of large bouts in each MOx signal, we first located the bouts on the band-pass filtered version of the signal. The lower and higher cut-off frequencies of the band-pass filter used here for all MOx recordings are 0.1 Hz and 1 Hz, respectively. A first-order band-pass filter has been employed here, in contrast to the first-order high-pass filter that was used to extract the high-frequency transients from the MOx signal in our previous work \cite{Rastogi2024}. The band-pass filter removes the high-frequency components from MOx recordings generated due to noise in the PID traces, in addition to the low-frequency components of the MOx sensor response that are a function of historical exposure to gases. The largest bout in the filtered signal from the time $t_{start}$ to $t_{end}$ was located by computing the maxima ($max_{value}$) and minima ($min_{value}$) of the signal within these time points, as well as their corresponding time of occurrences ($max_{t}$ and $min_{t}$). $min_{t}$ and $max_{t}$ represent the start and end time of this bout respectively. The value of timepoints $t_{start}$ and $t_{end}$ depends on the location of our interested peaks in the signal. In our case, we are looking at the largest bouts from 0 to 2s and from 2 to 5s for the plume traces in Table 1 and from 0 to 2s for the plume traces in Table 2. The slope of the bout can be computed using its amplitude ($max_{value}$-$min_{value}$) and duration ($max_{t}$-$min_{t}$) as follows: \par

\begin{equation}
    \label{eq_b_slope}
    b_{slope} = \frac{max_{value}-min_{value}}{max_{t}-min_{t}}
\end{equation}

\begin{figure}[h]
\includegraphics[width=\linewidth]{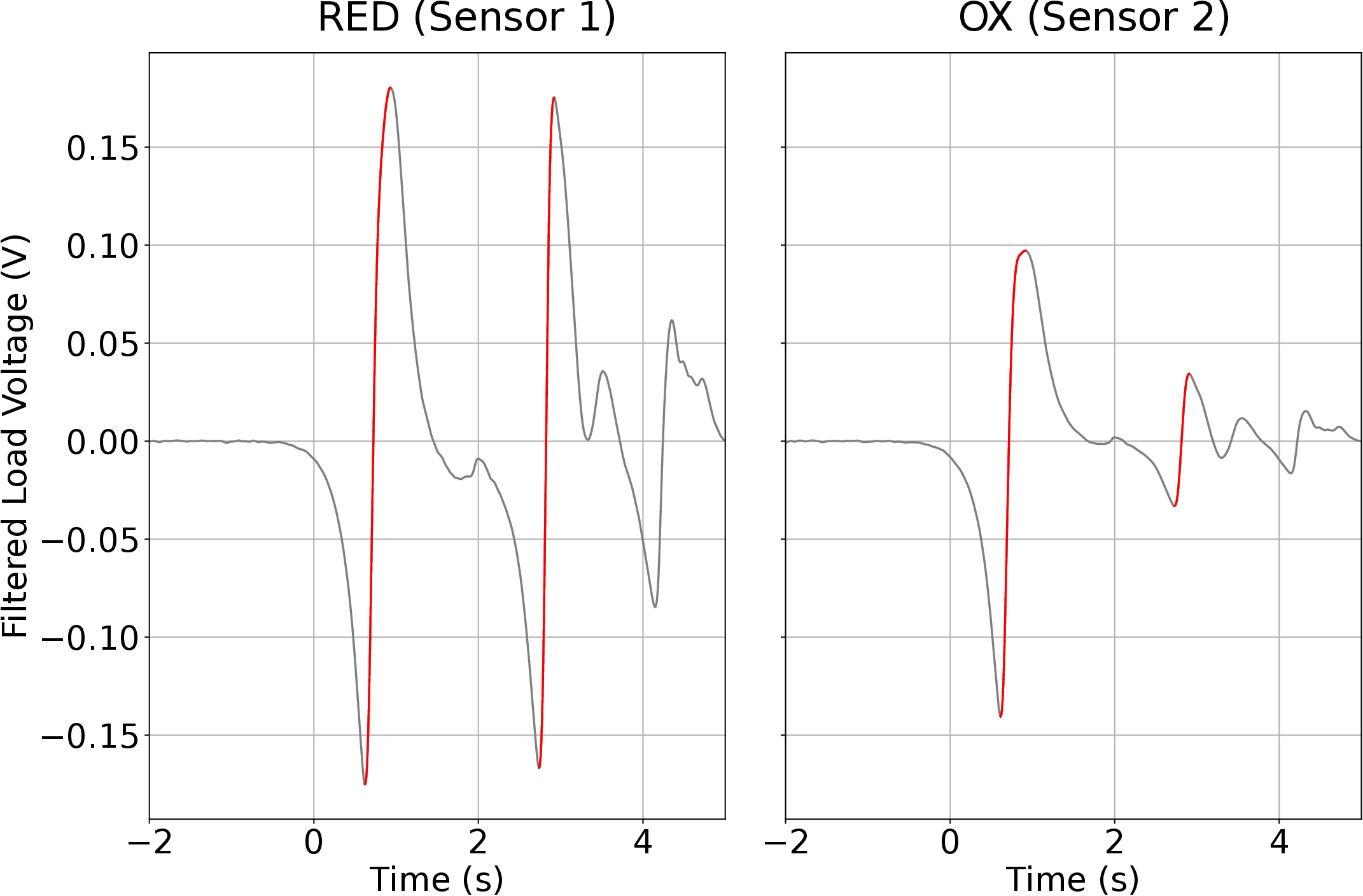}
\caption{Band-pass filtered MOx recordings for the plume trace Eu\textunderscore EB\textunderscore 2. The largest bouts in the duration 0-2s and 2-5s are highlighted in red.}
\label{fig:bouts_eb_eu_50_16_i}
\end{figure}

Figure 3 shows the band-pass filtered MOx recordings for the trace Eu\textunderscore EB\textunderscore 2 along with the largest bouts (highlighted in red) in the duration 0-2s and 2-5s. The bout in the duration 0-2s indicates the presence of Eu gas at the highest concentration level and the one in the duration 2-5s indicates the presence of EB gas at the same concentration level. \par

The bout slope was computed using (1) for each gas at two different peaks for all MOx recordings of plume traces mentioned in Table 1. The bar plots in Figure 4 show bout slopes for each gas at two different peaks in the plume traces of Table 1. It can be seen that bout slopes calculated for Peak 2 of each gas are different from those computed for Peak 1 of the same gas. The reason behind observed differences in bout slopes for Peak 1 and 2 could be that MOx recordings for Peak 2 do not start from baseline unlike in the case of Peak 1 (Figure 2). Therefore, bout slopes for Peak 2 might also encode information about the previous gas present in Peak 1. If this is the case, then Peak 1 is much more reliable than Peak 2 for gas identity recognition and concentration measurement in the available plume traces. \par

\begin{figure}[h]
\includegraphics[width=\linewidth]{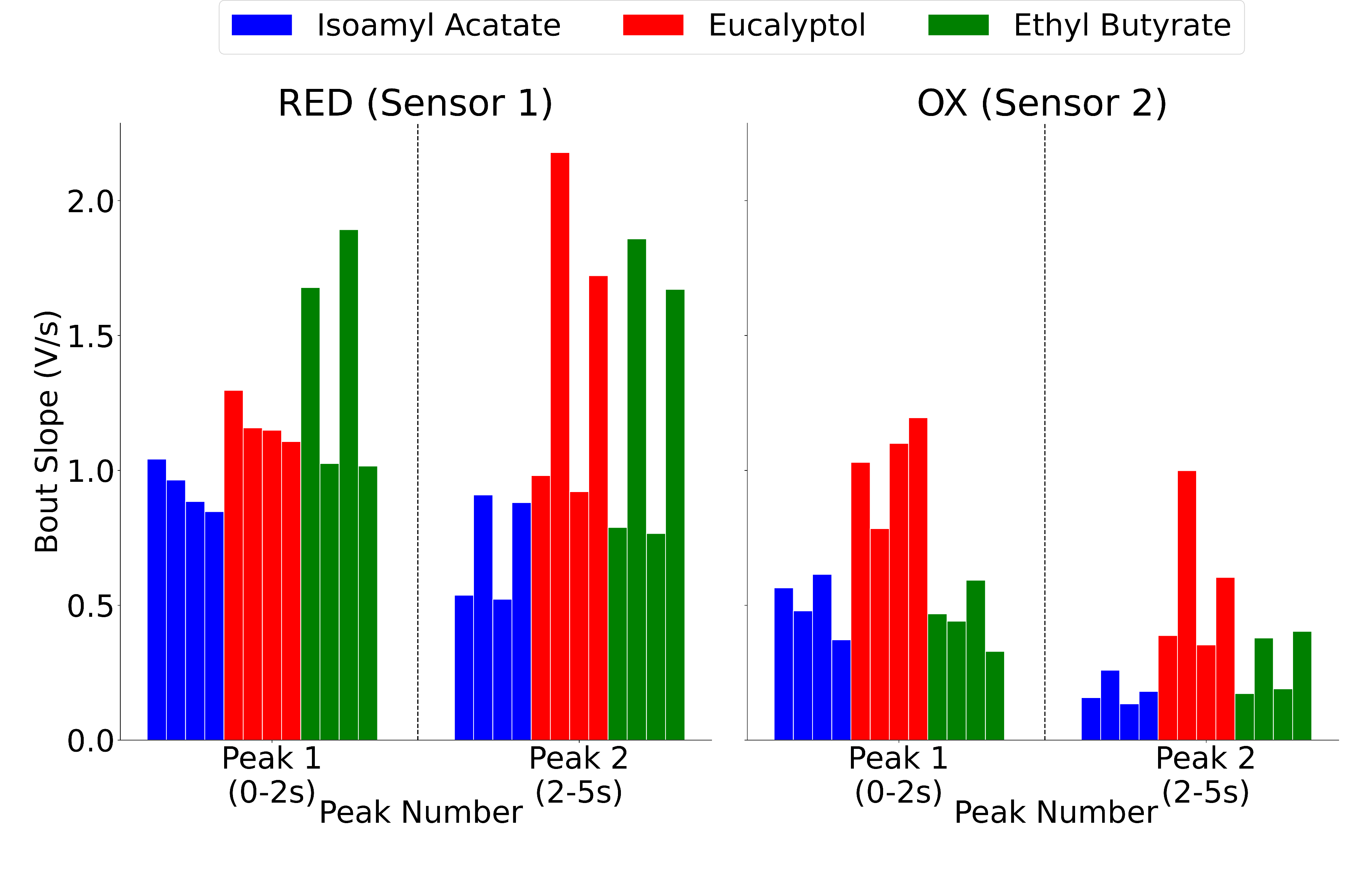}
\caption{Bar plots of bout slopes for all gases at two different peaks in the plume traces shown in Table 1.}
\label{fig:bar_plots_bout_slopes}
\end{figure}

It is evident from Figure 4 that the bout slopes obtained for Eu gas at Peak 1 look different from the other two gases and this difference is very significant for sensor 2. This means that we should be able to distinguish Eu gas from the other gases using some combinatorial bout slope output of two available sensors. This indicates that bout slopes can potentially be used to distinguish between different gases at Peak 1.  \par

\begin{figure}[H]
\centering
\includegraphics[width=\linewidth]{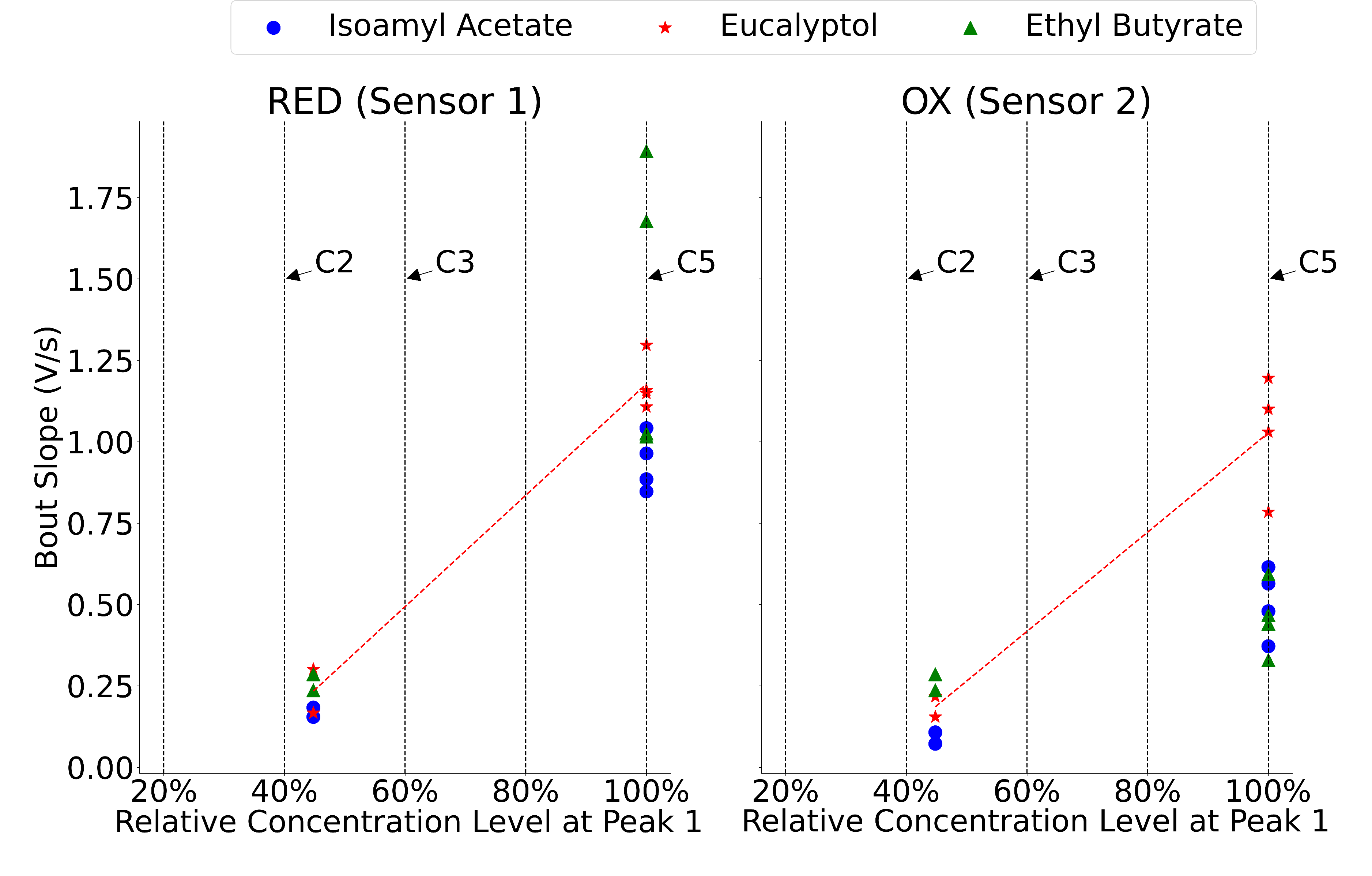}
\caption{Scatter plot of bout slopes for all gases at two different concentration levels at Peak 1. The x-axes represent different concentration levels in the form of percentages relative to the highest used gas concentration. The interpolated red dotted line depicts the rising trend for bout slopes of Eu gas with respect to gas concentration level.}
\label{fig:bouts_slopes_scatter_plot_peak_1_conc_levels}
\end{figure}

Figure 5 shows the scatter plots of bout slopes at Peak 1 for different gases at the lower concentration level (between C2 and C3) (Table 2) and at the highest concentration level (C5) (Table 1). It can be seen that bout slopes at the lower concentration level are lower than those at the highest concentration level. For sensor 1, the clustering of Eu gas bout slopes can be seen at the highest concentration level. For sensor 2, this clustering was observed at both concentration levels for Eu gas. The rising trend in the bout slope of Eu gas with respect to concentration levels is shown by the interpolated red dotted line on both sensors. This trend cannot be shown for the other two gases because bout slope points for other gases are not clustered and well separated like Eu gas at the highest concentration level on both sensors. However, the bout slopes of the other two gases form separate clusters at the lower concentration level for sensor 2. This indicates that some combinatorial bout slope output for the two sensors might be used for the gas identification at Peak 1.\par

With the available plume traces, it was observed that gases could potentially be distinguished from each other at the very first dominant peak using bout slopes and it seems that bout slopes can be used for gas identification at a specific concentration at the first peak in odor plumes. However, the dataset we have available for odor plumes is very small and it would not be wise to make a strong conclusion using this dataset. Therefore, we need to go back to the single gas pulse dataset to observe the performance of bout slopes for different gases and at different concentration levels. \par

\subsection{Bout Slope Analysis for Single Gas Pulse Dataset}

The MOx recordings obtained for the single gas pulse environment used for our previous study \cite{Rastogi2024} have only one big bout in the entire duration of a recording. These recordings were filtered using a first-order band-pass filter with lower and higher cut-off frequencies of 0.04 Hz and 1 Hz respectively. The bout was located by finding the time points of maxima and minima in these filtered recordings. The bout slope was computed using (1) for different trials of 3 gases at five different concentration levels. \par

\begin{figure}[H]
\centering
\includegraphics[width=\linewidth]{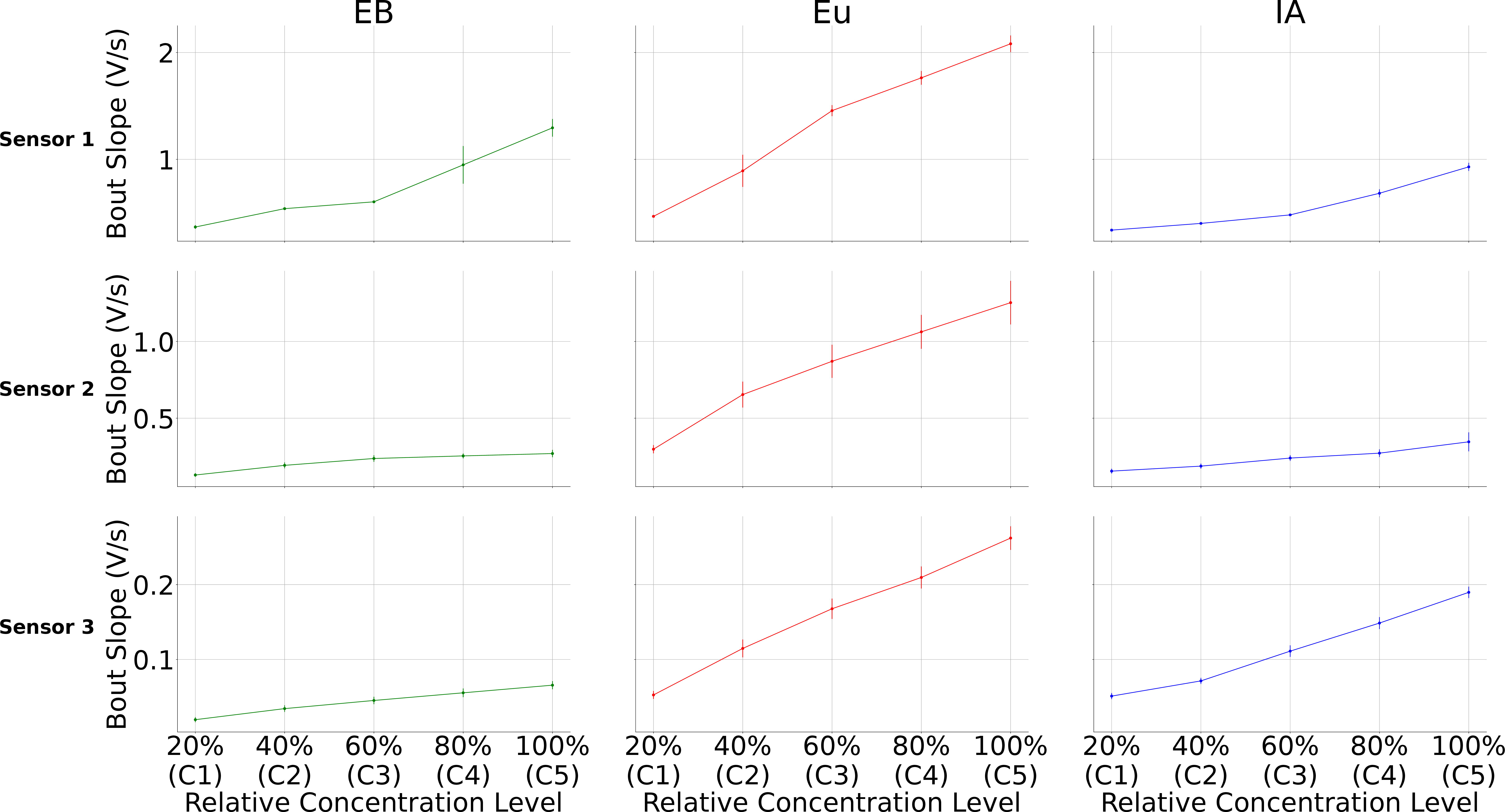}
\caption{Bout slope with respect to concentration level for single gas pulses. Dots and error bars represent the mean and standard deviation respectively over 20 trials.}
\label{fig:bouts_slope_gas_pulse_s123}
\end{figure}

Figure 6 shows the variation in bout slopes with respect to the concentration level in the single gas pulse environment for all 3 gases. It can be seen that bout slopes show a "strictly increasing" trend with respect to gas concentration level for all sensors. Also, the increment in bout slopes with the increase in concentration level is different for different gases for a given sensor. These observations together with the observations for odor plume traces show that the bout slope is a good indicator of gas at a particular concentration level both in a single gas pulse as well as for the first gas peak in a realistic environment. Therefore, we will modify the circuit design proposed in our previous study for the detection and measurement of bout slopes. The Skywater 130 PDK \cite{edwards2021introduction} has been used to design and simulate all circuits. \par

\section{Modified Circuit Design}

\subsection{Single Gas Pulse Environment}


Figure 7 shows the modified circuit design for the single gas pulse environment. The Change Detection stage is the same as in the previous study \cite{Rastogi2024}. The capacitance of the capacitor $C_{Ramp}$ in the Timer circuit has been reduced to 20 $\mu$F from 50 $\mu$F. This is because we don't have to wait for the integration operation for all MOx signals to complete now, unlike in the previous case. The high-pass filter and exposure measurement stages have now been replaced by the band-pass filter and slope detection stage respectively. The lower ($f_{L}$) and upper ($f_{H}$) cut-off frequencies of the band-pass filter are 0.04 Hz and 1 Hz respectively. The Slope Detection (SD) stage is a simple comparator that compares the band-pass filtered signal with an adjustable threshold. Whenever the band-pass filtered MOx signal crosses this threshold, a slope detection pulse is generated. Here, the fixed duration of the global trigger pulse represents the maximum duration for the slope measurement. Therefore, we need to avoid the slope detection pulse being ON when the global trigger pulse is OFF. This is ensured by the AND gates G4, G5, and G6 that perform AND operations on the global trigger pulse $Q_{out}$ and on the SD pulses generated by comparators U1, U2, and U3 to generate new slope detection pulses $Out_{S1}$,$Out_{S2}$ and $Out_{S3}$.   \par

\begin{figure}[H]
\centering
\includegraphics[width=\linewidth]
{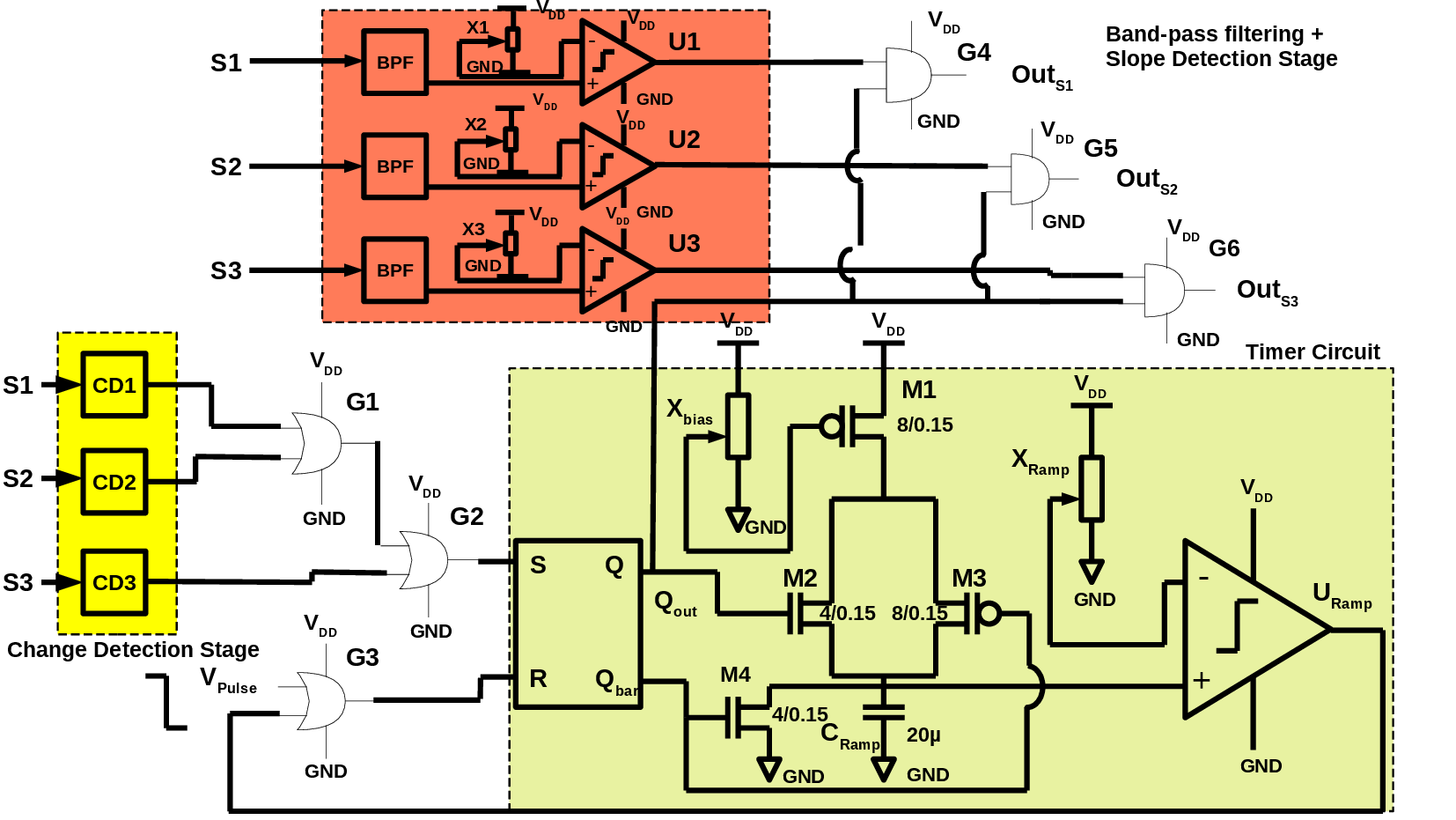}
\caption{The modified analog front-end design for single gas pulse environment.}
\label{fig:ckt_design_slope_detection}
\end{figure}

\subsection{Plume Environment}

After modifying the circuit for the single gas pulse environment, a similar circuit for the plume environment was designed. Figure 8 shows the circuit design for MOx signals obtained using gas plume traces depicted in Figure 1. As mentioned before, recordings of sensors 1 and 2 only have been used for the plume environment. All the circuit parameters are the same as in the single gas pulse environment except the cut-off frequencies of band-pass filter $BPF_{plume}$ and the resistor values obtained using potentiometers X1 and X2. For the case of gas plumes, the lower ($f_{L}$) and upper ($f_{H}$) cut-off frequencies of the band-pass filter are 0.4 Hz and 1 Hz, respectively. 
The lower cut-off frequency of the band-pass filter has been chosen to be higher than the one used for the analysis of bout slopes (which was 0.1 Hz) to ensure a faster decay of the MOx signal after the release of gas. When the MOx signal decays faster, the SD pulse will be off before the resetting of $Q_{out}$ pulse, eliminating the need for AND gates. Therefore, this setting makes this circuit more compact.\par 

\begin{figure}[H]
\centering
\includegraphics[width=\linewidth]{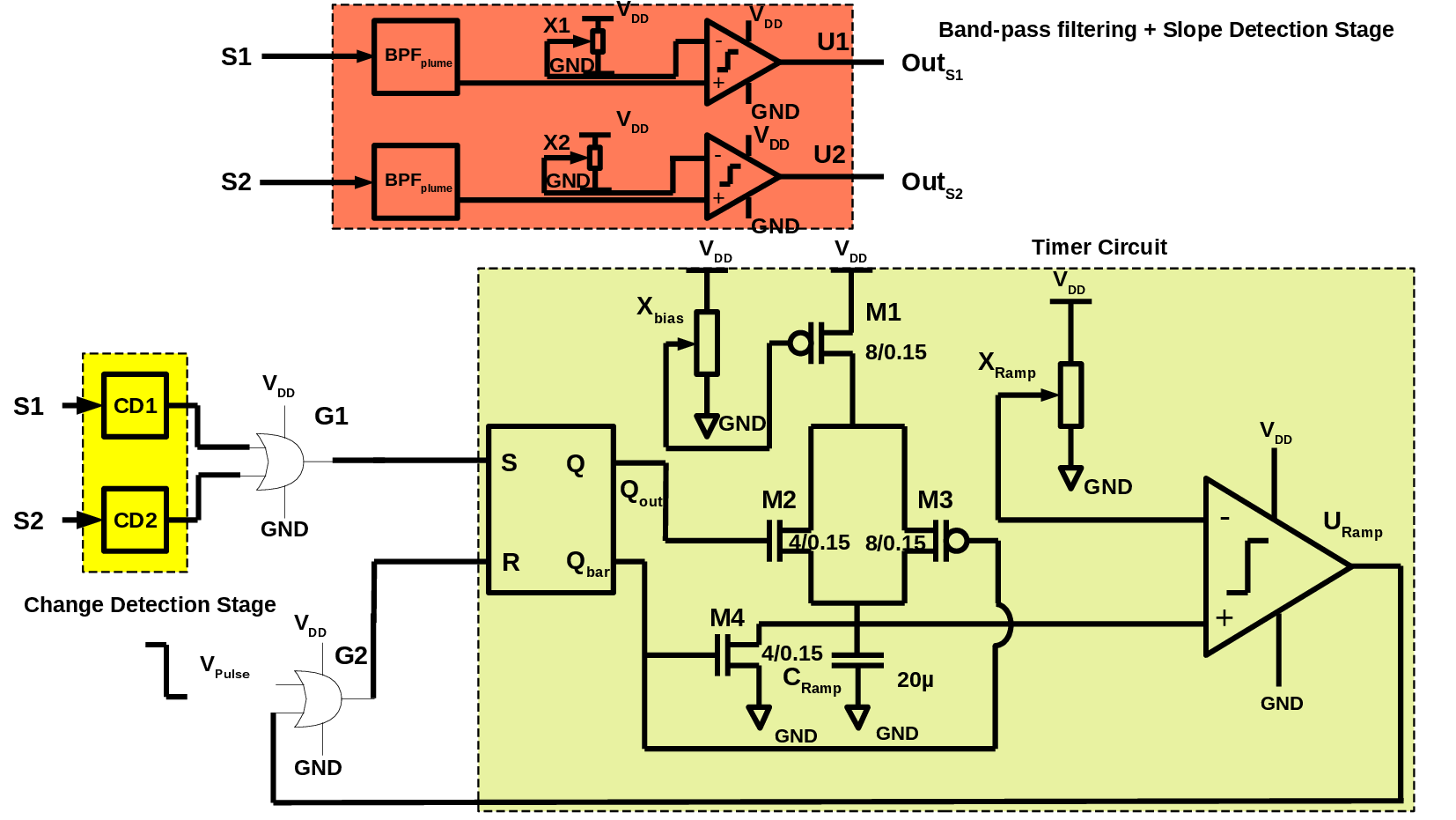}
\caption{The analog front-end design for the plume environment.}
\label{fig:ckt_design_slope_detection_plume}
\end{figure}

\section{Results}

\subsection{Single Gas Pulse Environment}

The proposed modified circuit was simulated using the MOx data for three gas pulses at five different concentration levels and the time difference between the SD pulse ($Out_{S1}$, $Out_{S2}$ and $Out_{S3}$) and the global trigger pulse ($Q_{out}$) was obtained for 3 sensors in the array. Figure 9 shows the output of the modified circuit for EB gas at concentration level 1. It can be seen that SD pulses for all 3 sensors detect the rising edges of band-pass filtered signal on these sensors. SD pulses are on as long as the $Q_{out}$ pulse is on and are off when the $Q_{out}$ pulse is off. Thus, the slope detection is confined by the duration of $Q_{out}$ pulse. \par

\begin{figure}[H]
\centering
\includegraphics[width=\linewidth]
{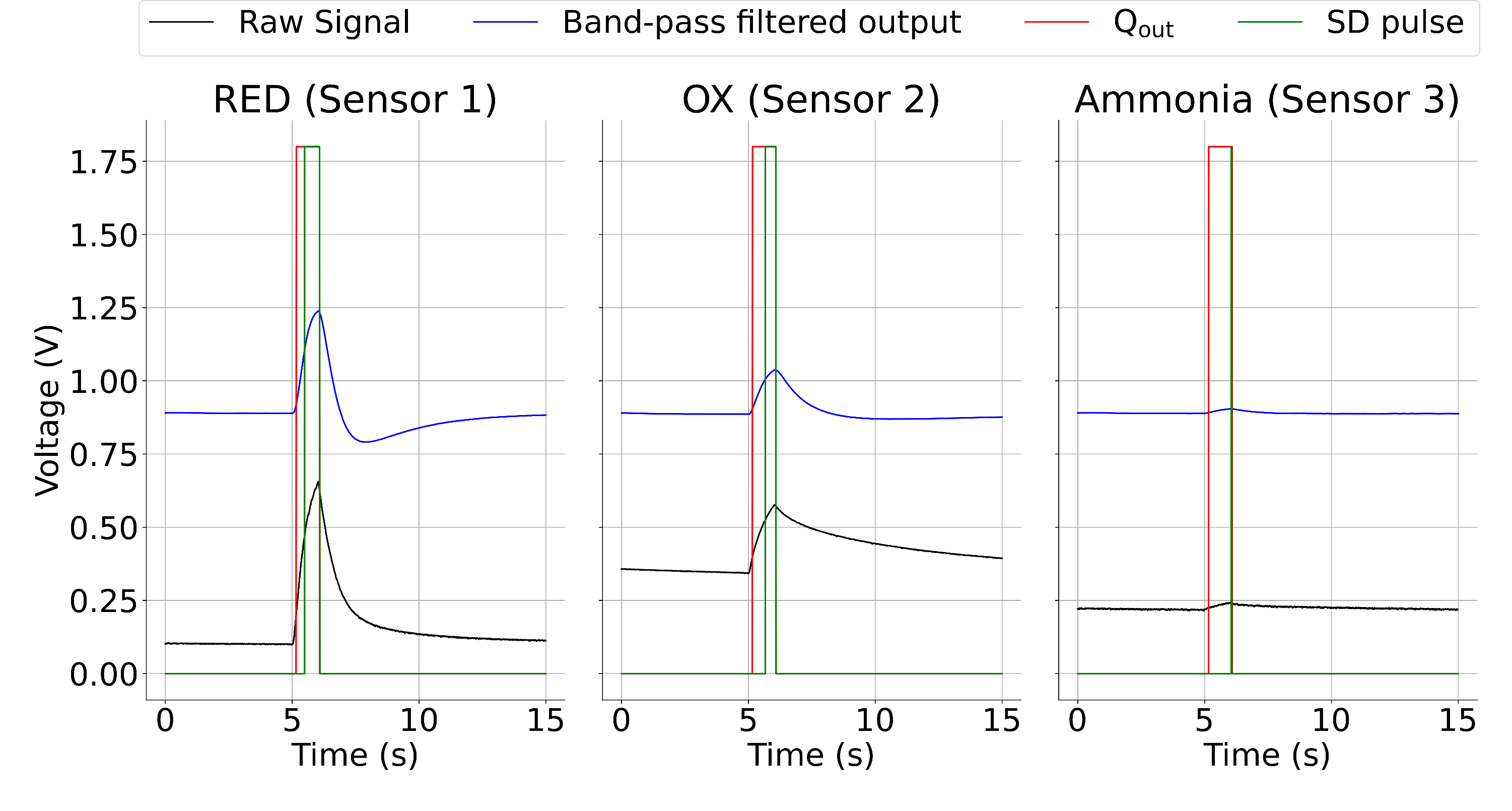}
\caption{Modified circuit output for Ethyl Butyrate gas at Concentration Level 1}
\label{fig:modified_ckt_output_EB_single_pulse}
\end{figure}

\begin{figure}[H]
\includegraphics[width=\linewidth]
{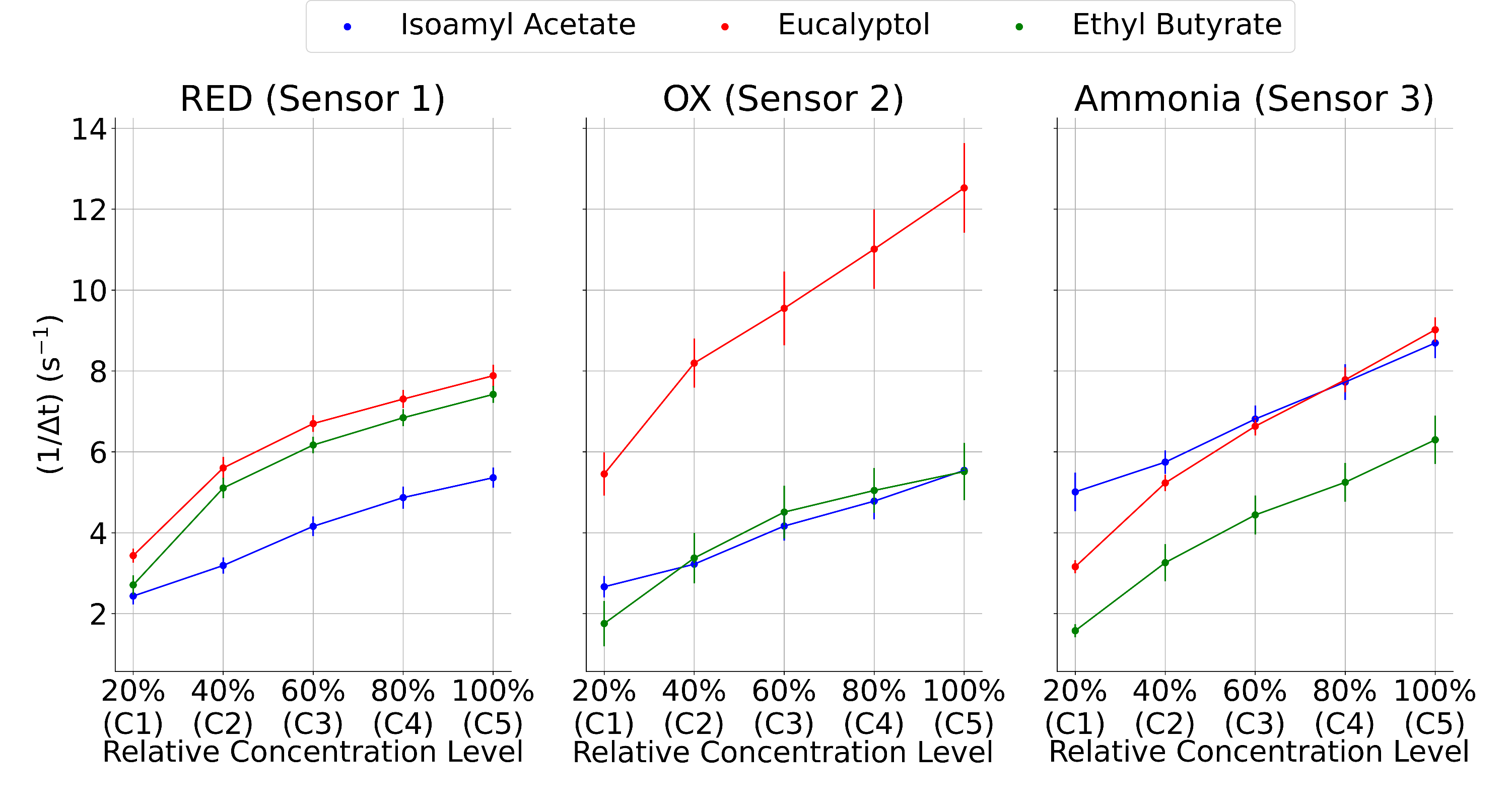}
\caption{Inverse of the time difference between $Q_{out}$ and SD pulse activation with respect to concentration level for 3 sensors of MiCS-6814. The x-axes represent different concentration levels in the form of percentages relative to the highest used gas concentration. Dots and error bars represent the mean and standard deviation across 20 trials respectively. For one trial in sensor 2 and another trial in sensor 3, the SD pulse was not observed for EB gas and therefore these two trials for these sensors are discarded. }
\label{fig:conc_plots_slope_detection}
\end{figure}

Figure 10 shows the plot of the inverse of the mean time difference between $Q_{out}$ and SD pulse activation over all trials for 3 sensors of MiCS-6814. It can be seen that the trend followed by the time difference between $Q_{out}$ and SD pulse with respect to the gas concentration level is similar to that followed by the time difference between $Q_{out}$ and EM pulse observed in our previous study \cite{Rastogi2024}. However, there are some obvious differences between these two results. The slopes of the curves for all gases and for all 3 sensors are different from the ones observed in the previous case \cite{Rastogi2024}. Also, the sizes of error bars for slope detection are bigger than those observed for the exposure measurement. For a single gas pulse environment with a controlled airflow, the slope of the signal is much more prone to variability between different trials than the area under the MOx response curve. For a gas pulse of constant amplitude embedded in controlled airflow conditions, the exposure measurement is therefore much more robust to noise than slope detection. In a turbulent environment, however, the area under the MOx response curve keeps on changing for different trials of the same gas at a given concentration level (Figure 2) because of temporal variations in plume dynamics. Therefore, the slope of the signal exhibits higher robustness to noise in such environments.



\subsection{Plume Environment}
The circuit was simulated for all the plume recordings shown in Table 1 and the time difference between $Q_{out}$ and SD pulse was computed for the first two prominent peaks. Figure 11 shows the circuit output of MOx recordings for the plume EB\textunderscore IA\textunderscore 2. It can be seen that the circuit detects the rising edges of the first two prominent peaks in the band-pass filtered MOx signal. \par
The bar plots in Figure 12 show the inverse of time differences between $Q_{out}$ and SD pulse for 3 tested gases obtained for the first two prominent peaks on both sensors. For Peak 1, the circuit output of EB gas is slightly different from the other two gases in the case of sensor 1. For sensor 2, the circuit output of Eu gas is very different from the other two gases at Peak 1. 
Gas recognition using Peak 2 is not straightforward as the circuit output at the shape of this peak depends on the history of exposure. \par

\begin{figure}[h]
\includegraphics[width=\linewidth]{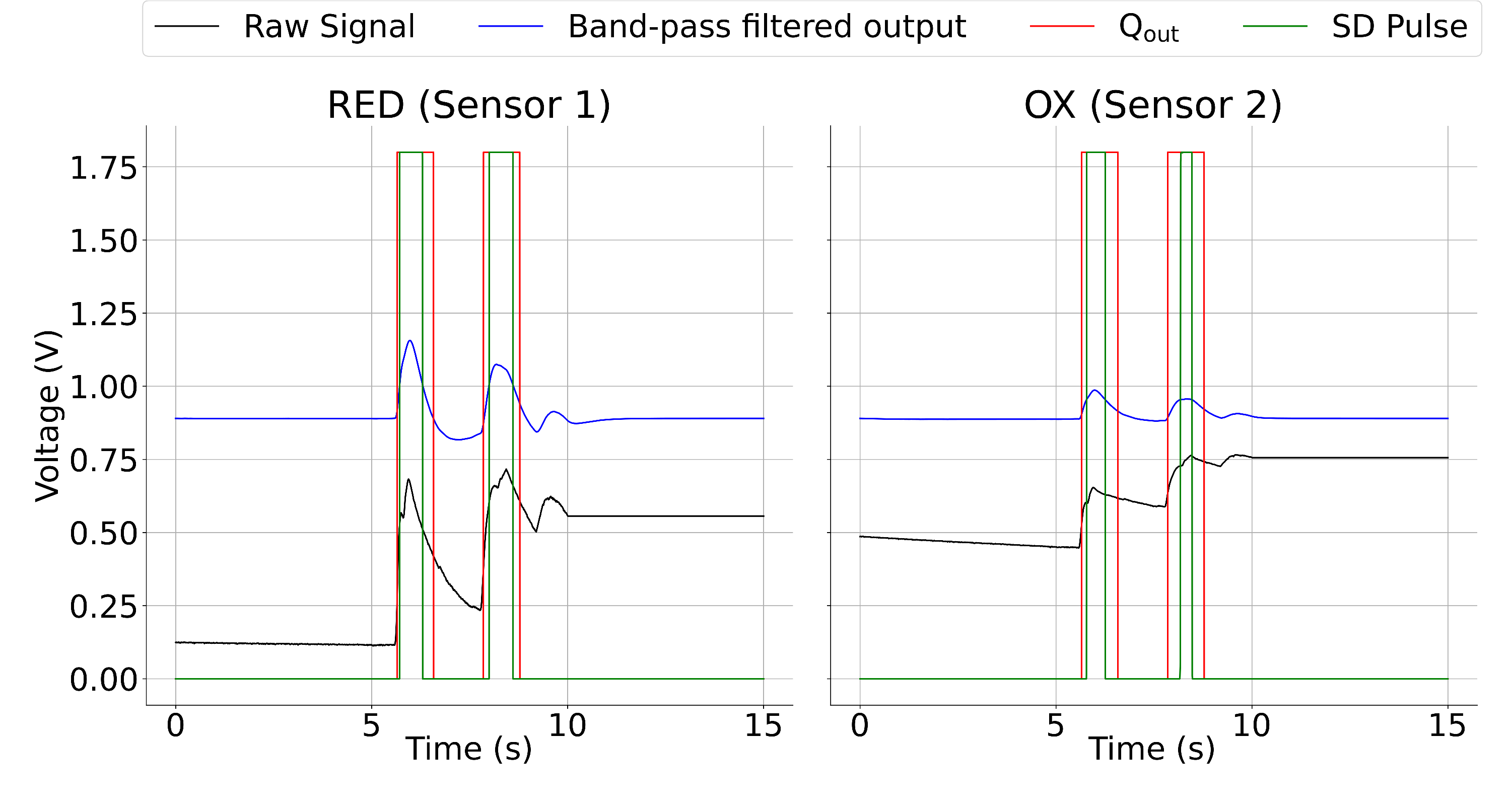}
\caption{Circuit output for MOx recordings of plume trace EB\textunderscore IA\textunderscore50\textunderscore16}
\label{fig:ckt_output_EB_IA_50_16}
\end{figure}

\begin{figure}[H]
\centering
\includegraphics[width=\linewidth]{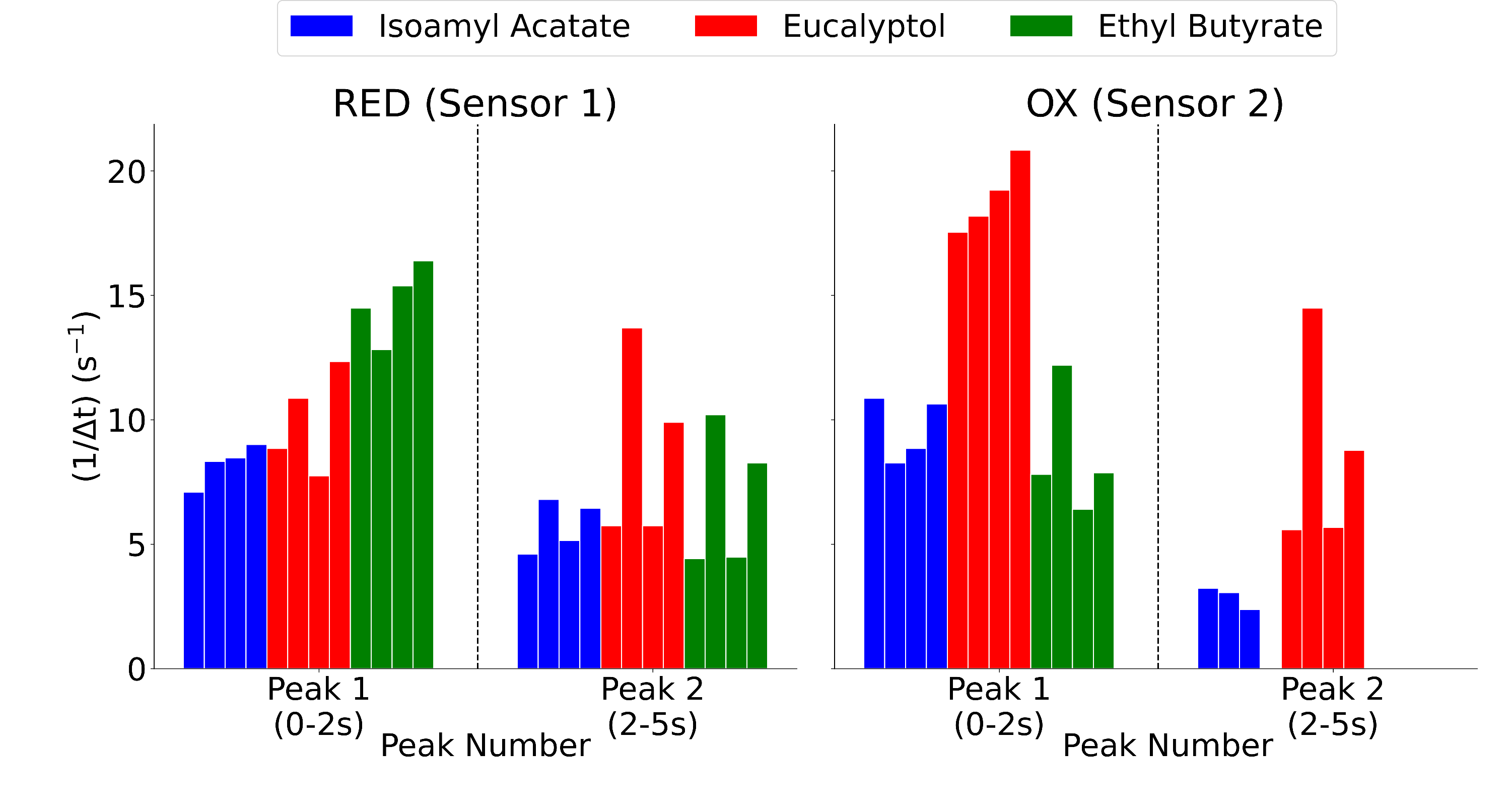}
\caption{Bar plots of the inverse of time difference computed for two different peaks in the plume recordings of Table 1}
\label{fig:b_plot_t_diff_inv_peak_1_2}
\end{figure}

Figure 13 shows the scatter plots of the inverse of the time difference computed for two different concentration levels at Peak 1 using plume recordings of Tables 1 and 2.  It is not possible to distinguish between different gases at the lower concentration level using the circuit output of sensor 1. However, Eu gas can be distinguished from the other two gases at this concentration level using the circuit output of sensor 2. The observations for EB gas at the highest concentration level form a separate cluster for sensor 1 and those for Eu and IA gas at the same concentration level form clusters for sensor 2. Therefore, the rising trend in the circuit output with respect to gas concentration for the EB gas is shown by the interpolated green dotted line on sensor 1 and those for Eu and IA gas are shown by interpolated red and blue dotted lines respectively on sensor 2. \par

\begin{figure}[H]
\centering
\includegraphics[width=\linewidth]{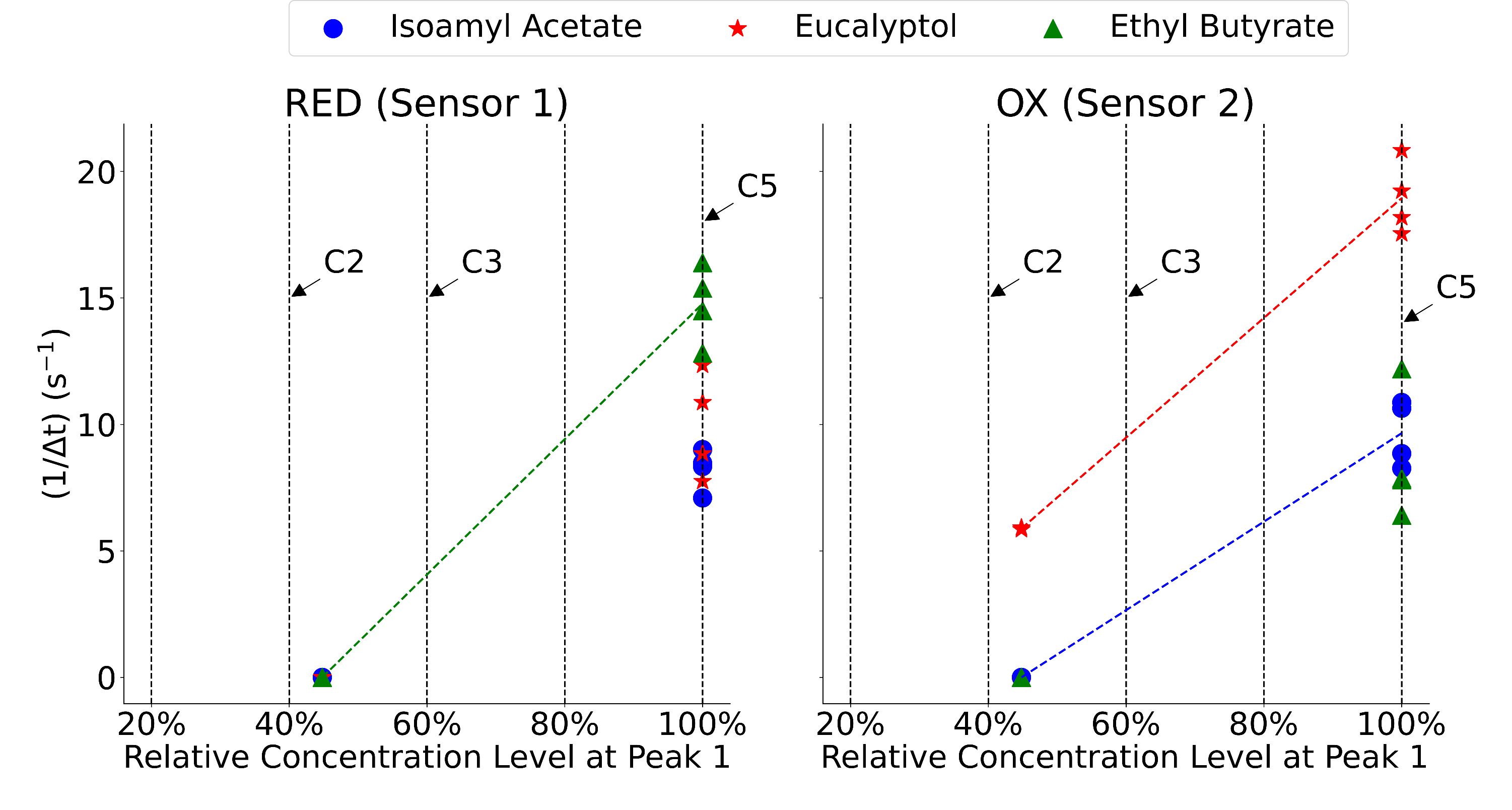}
\caption{Scatter plot of the inverse of time differences obtained on two sensors at different concentration levels at Peak 1. The x-axes represent different concentration levels in the form of percentages relative to the highest used gas concentration. The rising trend with respect to gas concentration level for EB gas on sensor 1 is shown by the interpolated green dotted line. Similar trends for IA and Eu gas on sensor 2 are shown by interpolated blue and red dotted lines respectively.}
\label{fig:s_plot_t_diff_inv_peak_1_conc_levels}
\end{figure}

The scatter plot in Figure 14 shows the sum of circuit outputs obtained on sensors 1 and 2 for 3 gases at Peak 1 for both concentration levels. The clustering can now be observed in the summed output for all 3 gases at the highest concentration level. It is now easier to observe the rising trend with respect to gas concentration for all gases (shown by interpolated dotted lines). At the lower concentration level, only Eu gas can be distinguished from the rest of the other gases but all gases can be distinguished from each other at the highest concentration level using the summed output. This shows that the combinatorial output of the circuit for all sensors is much more powerful in the plume environment than looking at the circuit output of individual sensors separately for gas recognition, especially for higher gas concentration levels. However, it should be noted that this combinatorial circuit output can be used either to recognize the gas at a known concentration level or to estimate the concentration level of a known gas. Observing the output of multiple sensors of the e-nose separately provides us with multiple dimensions depending on the number of sensors used which helps us to separate gas identity from its concentration level. By combining the output of multiple sensors into one, we have to give up on this dimensionality which allows us to decode only one out of these two possible variables. In the available dataset, the different gases at the first prominent peak have only two possible concentration levels. Therefore, we can test our circuit only for these two concentration levels. Further experiments are needed to obtain the plume data at multiple gas concentration levels to validate these observations for different concentrations in a realistic scenario.  


\begin{figure}[h]
\includegraphics[width=\linewidth]{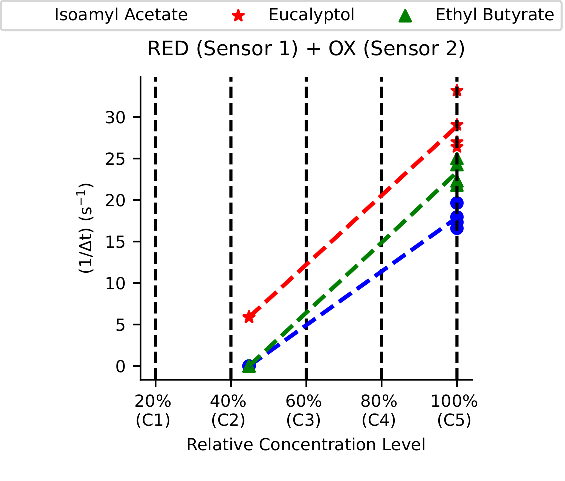}
\caption{Scatter plot of inverse of time difference computed for two different concentration levels at Peak 1}
\label{fig:s_plot_t_diff_inv_peak_1_conc_levels_s1+2}
\end{figure}

\section{Conclusion and Discussion}
In this work, 
we have introduced a sensory front-end design tailored for integration with an electronic nose, enabling reliable gas detection and concentration estimation in turbulent plume environments. This is achieved through cascaded analog event generation, encoding the gas concentration in the time difference between two events.

Currently, the proposed circuit detects prominent gas peaks reliably, yet faces challenges in gas detection and concentration estimation when peaks are not well separated.
In the available plume data, the two consecutive prominent peaks occur within two seconds or less. As this is shorter than the full recovery period of the used MOx sensors,  
the circuit output generated for the second peak 
is not as reliable as the one generated for the first. 
To verify whether the second peak can be as reliable as the first one in plume traces where the peaks are well separated, further data collection campaigns will be needed. 
Alternatively, further research should be done on processing the signal, potentially enhancing the performance under challenging conditions.
\par

Further, the proposed circuit detects only lower frequency dynamics of plumes like whiffs, blanks, and plume onsets \cite{Celani2014}. Plume dynamics at higher frequencies (e.g.: clumps \cite{Celani2014}) cannot be captured by this circuit. This is because the duration of change detection is fixed by the SR latch in the Timer circuit and any new change within this duration cannot be detected. To detect higher-order temporal features of the plume, the capacitance $C_{ramp}$ of the ramp-generating circuit could be reduced. Doing so, however, may cause the SR latch to enter an unpredictable state, potentially leading to unreliable measurements. \par

Lewis et al. \cite{Lewis2024} found that the majority of spiking activity of the mouse olfactory bulb (OB) does not encode the full range of odor concentration dynamics and high-frequency fluctuations in the plume. Instead, only relatively low-level temporal features like whiffs, blanks, plume onset, and plume offset are represented in the OB. Therefore, it can be said that our proposed neuromorphic sensory front-end encodes the plume dynamics similar to a mammalian OB.\par

The proposed circuits in this as well as in the previous study \cite{Rastogi2024}, however, do not replicate the exact spiking output of mitral and tufted cells. These circuits encode the gas concentration in a way similar to these principal output neurons in the OB which allows us to understand the mechanism of OB circuits for the specific task of odor concentration encoding in a simplified way. Tasks such as odor plume navigation require sustained spiking activities of mitral and tufted cells \cite{Lewis2024}\cite{Patterson2013} which are responsible for the behavioral switch of an animal during navigation. If we want to build a robot in the future that can navigate towards an odor source in a plume environment with behavioral strategies similar to those of an animal, our circuit might not perform well in this scenario because it generates only one event each from the CD circuit (tufted cell model) and the SD circuit (mitral cell model). A more biologically plausible electronic model of parallel pathways in the OB might be useful for such tasks, similar to models developed for other sensory modalities \cite{Kameda2003}\cite{Philip2024}.

\bibliographystyle{ieeetr}

\bibliography{ref}

\vspace{12pt}

\section*{Author contributions statement}
\textbf{Shavika Rastogi:} Conceptualization, Methodology, Formal analysis, Software, Investigation, Writing - Original draft preparation. \textbf{Nik Dennler:} Data curation, Writing - Reviewing and Editing. \textbf{Michael Schmuker:} Supervision, Funding acquisition, Writing - Reviewing and Editing. \textbf{André van Schaik:} Conceptualization, Supervision, Funding acquisition, Writing - Reviewing and Editing.

\end{document}